%% file: acl2017.tex
\documentclass[11pt,a4paper]{article}
\usepackage{authblk}
\usepackage[hyperref]{acl2017}
\usepackage{times}
\usepackage{multirow}
\usepackage{comment}
\usepackage{amssymb}
\usepackage{amsmath}
\usepackage{booktabs}
\usepackage{graphicx}
\usepackage{setspace}
\usepackage{caption}

\usepackage[noend]{myalgorithmic}
\usepackage{myalgorithm}

\usepackage{pgfplots}

\usepackage{url}

\DeclareMathOperator*{\argmax}{\arg\!\max}

\renewcommand{\algorithmicrequire}{\textbf{Input:}}
\renewcommand{\algorithmicensure}{\textbf{Output:}}
\renewcommand{\algorithmicforall}{\textbf{for each}}
\renewcommand{\algorithmicdo}{}

\aclfinalcopy 

\title{Towards a Seamless Integration of Word Senses \\ into Downstream NLP Applications}

\author[2]{\bf Mohammad Taher Pilehvar}
\author[1]{\bf Jose Camacho-Collados}
\author[1]{\\ \bf Roberto Navigli}
\author[2]{\bf Nigel Collier}

\affil[1]{Department of Computer Science, Sapienza University of Rome}
\affil[2]{Department of Theoretical and Applied Linguistics, University of Cambridge}

\affil[1]{\tt \{collados,navigli\}@di.uniroma1.it}
\affil[2]{\tt  \{mp792,nhc30\}@cam.ac.uk}

\date{}

\begin{document}





\maketitle

\begin{abstract}
  Lexical ambiguity can impede NLP systems from accurate understanding of semantics. Despite its potential benefits, the integration of sense-level information into NLP systems has remained understudied. By incorporating a novel disambiguation algorithm into a state-of-the-art classification model, we create a pipeline to integrate sense-level information into downstream NLP applications. We show that a simple disambiguation of the input text can lead to consistent performance improvement on multiple topic categorization and polarity detection datasets, particularly when the fine granularity of the underlying sense inventory is reduced and the document is sufficiently large. Our results also point to the need for sense representation research to focus more on \textit{in vivo} evaluations which target the performance in downstream NLP applications rather than artificial benchmarks. 
 
\end{abstract}


\section{Introduction}

As a general trend, most current Natural Language Processing (NLP) systems
function at the word level, i.e. individual words constitute the most fine-grained meaning-bearing elements of their input.
The word level functionality can affect the performance of these systems in two ways:
(1) 
it can hamper their efficiency in handling words that are not encountered frequently during training, such as multiwords, inflections and derivations, and 
(2)
it can restrict their semantic understanding to the level of words, with all their ambiguities, and thereby prevent accurate capture of the intended meanings.

The first issue has recently been alleviated by techniques that aim to boost the generalisation power of NLP systems by resorting to sub-word or character-level information \cite{ballesterosimproved,kim2015character}. 
The second limitation, however, has not yet been studied sufficiently.
A reasonable way to handle word ambiguity, and hence to tackle the second issue, is to \textit{semantify} the input text: transform it from its surface-level semantics to the deeper level of word senses, i.e. their intended meanings.
We take a step in this direction by designing a pipeline that enables seamless integration of word senses into downstream NLP applications, while benefiting from knowledge extracted from semantic networks. To this end, we propose a quick graph-based Word Sense Disambiguation (WSD) algorithm which allows high confidence disambiguation of words without much computation overload on the system. 
We evaluate the pipeline in two downstream NLP applications: polarity detection and topic categorization. Specifically, we use a classification model based on Convolutional Neural Networks which has been shown to be very effective in various text classification tasks \cite{kalchbrenner2014convolutional,kim2014convolutional,johnson2015effective,tang2015document,XiaoCho:2016}. 
We show that a simple disambiguation of input can lead to performance improvement of a state-of-the-art text classification system on multiple datasets, particularly for long inputs and when the granularity of the sense inventory is reduced. Our pipeline is quite flexible and modular, as it permits the integration of different WSD and sense representation techniques.





\section{Motivation}

\begin{figure}[t!]
\begin{center}
	\includegraphics[trim = 0mm 10mm 0mm 0mm,,scale=0.26]{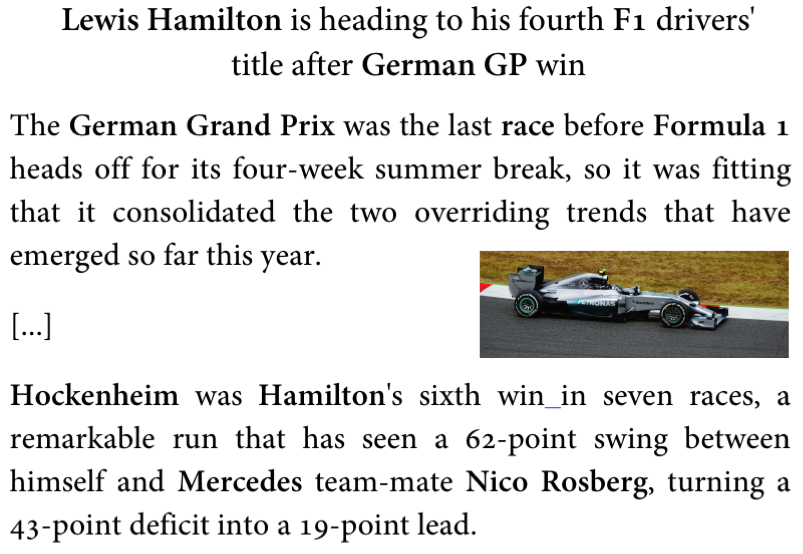}
\end{center}
    \caption{Excerpt of a news article from the BBC.}
    \label{fig:intuition}
\end{figure}

With the help of an example news article from the BBC, shown in Figure \ref{fig:intuition}, we highlight some of the potential deficiencies of word-based models.

\paragraph{Ambiguity.} 
Language is inherently ambiguous.
For instance, \textit{Mercedes}, \textit{race}, \textit{Hamilton} and \textit{Formula} can refer to several different entities or meanings.
Current neural models have managed to successfully represent complex semantic associations by effectively analyzing large amounts of data. 
However, the word-level functionality of these systems is still a barrier to the depth of their natural language understanding.
Our proposal is particularly tailored towards addressing this issue. 


\paragraph{Multiword expressions (MWE).}
 MWE are lexical units made up of two or more words which are idiosyncratic in nature \cite{sag2002multiword}, e.g, \textit{Lewis Hamilton}, \textit{Nico Rosberg} and \textit{Formula 1}.
 Most existing word-based models ignore the interdependency between MWE's subunits and treat them as individual units.
Handling MWE has been a long-standing problem in NLP and has recently received a considerable amount of interest \cite{tsvetkov2014identification,salehi2015word}. 
Our pipeline facilitates this goal. 


\paragraph{Co-reference.} Co-reference resolution of concepts and entities is not explicitly tackled by our approach. However, thanks to the fact that words that refer to the same meaning in context, e.g., \textit{Formula 1}-\textit{F1} or \textit{German Grand Prix}-\textit{German GP}-\textit{Hockenheim}, are all disambiguated to the same concept, the co-reference issue is also partly addressed by our pipeline.

     

\section{Disambiguation Algorithm}
\label{disambiguation}

Our proposal relies on a seamless integration of word senses in word-based systems.
The goal is to semantify the text prior to its being fed into the system by transforming its individual units from word surface form to the deeper level of word senses.
The semantification step is mainly tailored towards resolving ambiguities, but it brings about other advantages mentioned in the previous section.
The aim is to provide the system with an input of reduced ambiguity which can facilitate its decision making.


To this end, we developed a simple graph-based joint disambiguation and entity linking algorithm which can take any arbitrary semantic network as input. 
The gist of our disambiguation technique lies in its speed and scalability.
Conventional knowledge-based disambiguation systems \cite{hoffart2012kore,agirre2014random,Moroetal:14tacl,ling2015design,PilehvarNavigli:2014b} often rely on computationally expensive graph algorithms, which limits their application to on-the-fly processing of large number of text documents, as is the case in our experiments.
Moreover, unlike supervised WSD and entity linking techniques \cite{ZhongNg:2010,cheng2013relational,melamud2016context2vec,limsopatham-collier:2016}, our algorithm relies only on semantic networks and does not require any sense-annotated data, which is limited to English and almost non-existent for other languages.


\begin{algorithm}[t!]
\caption{Disambiguation algorithm} 
\label{alg1}                         
\small
\begin{algorithmic}[1]
  \REQUIRE Input text $T$ and semantic network $N$   
  \ENSURE Set of disambiguated senses $\hat{S}$
  
    \STATE{Graph representation of $T$: $(S,E) \gets \mathrm{getGraph}(T, N)$}
    \STATE {$\hat{S} \gets \emptyset$} \label{line:beginning}  
    \FORALL {iteration $i \in \{1,...,len(T)\}$ } \label{line:mainFor11}       
    
        \STATE {$\hat{s}=\argmax_{s \in S} {|\{(s,s') \in E : s' \in S\}|}$}
        
        \STATE {$\mathrm{maxDeg}= {|\{(\hat{s},s') \in E : s' \in S\}|}$}
        
        \IF{$\mathrm{maxDeg} < \theta |S| \mathbin{/}{100} $} 
            \STATE{\textit{break}}
        \ELSE
            \STATE{$\hat{S} \gets \hat{S} \cup \{\hat{s}\}$}
            \STATE{$E \gets E \setminus \{ (s,s') : s \lor s' \in \mathrm{getLex}(\hat{s}) \}$}     \label{line:removal}        
        \ENDIF

    \ENDFOR \label{line:endfor}

    \RETURN Disambiguation output $\hat{S}$ \label{line:return1}

\end{algorithmic}
\end{algorithm}

Algorithm \ref{alg1} shows our procedure for disambiguating an input document $T$.
First, we retrieve from our semantic network the list of candidate senses\footnote{As defined in the underlying sense inventory, up to trigrams. We used Stanford CoreNLP \cite{manning-EtAl:2014:P14-5} for tokenization, Part-of-Speech (PoS) tagging and lemmatization.} for each content word, as well as semantic relationships among them. 
As a result, we obtain a graph representation $(S,E)$ of the input text, where $S$ is the set of candidate senses and $E$ is the set of edges among different senses in $S$. The graph is, in fact, a small sub-graph of the input semantic network, $N$.
Our algorithm then selects the best candidates iteratively. 
In each iteration, the candidate sense that has the highest graph degree $\mathrm{maxDeg}$ is chosen as the winning sense:
\begin{equation}
\mathrm{maxDeg}=\max_{s \in S} {|\{(s,s') \in E : s' \in S\}|}
\end{equation}

After each iteration, when a candidate sense $\hat{s}$ is selected, all the possible candidate senses of the corresponding word (i.e. $\mathrm{getLex}(\hat{s})$) are removed from $E$ (line  \ref{line:removal} in the algorithm).

\begin{figure}[t!]
\begin{center}
	\includegraphics[trim = 0mm 5mm 0mm 0mm,scale=0.26]{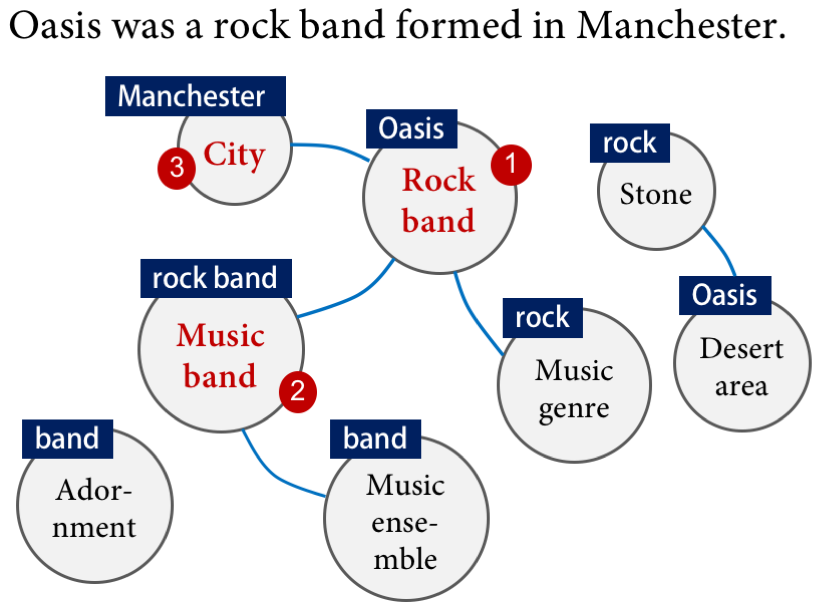}
\end{center}
    \caption{Simplified graph-based representation of a sample sentence.}
    \label{fig:oasis}
\end{figure}

Figure \ref{fig:oasis} shows a simplified version of the graph for a sample sentence.
The algorithm would disambiguate the content words in this sentence as follows. 
It first associates \textit{Oasis} with its \textit{rock band} sense, since its corresponding node has the highest degree, i.e. 3. 
On the basis of this, the \textit{desert} sense of \textit{Oasis} and its link to the \textit{stone} sense of \textit{rock} are removed from the graph. 
In the second iteration, \textit{rock band} is disambiguated as \textit{music band} given that its degree is 2.\footnote{For bigrams and trigrams whose individual words might also be disambiguated (such as \textit{rock} and \textit{band} in \textit{rock band}), the longest unit has the highest priority (i.e. \textit{rock band}).}
Finally, \textit{Manchester} is associated with its \textit{city} sense (with a degree of 1).

In order to enable disambiguating at different confidence levels, we introduce a threshold $\theta$ which determines the stopping criterion of the algorithm.
Iteration continues until the following condition is fulfilled: $\mathrm{maxDeg}<\theta |S|\mathbin{/}{100}$. 
This ensures that the system will only disambiguate those words for which it has a high confidence and backs off to the word form otherwise, avoiding the introduction of unwanted noise in the data for uncertain cases or for word senses that are not defined in the inventory.


\section{Classification Model}
\label{model}

\begin{figure}[t!]
\begin{center}
	\includegraphics[trim = 0mm 5mm 0mm 0mm,scale=0.27]{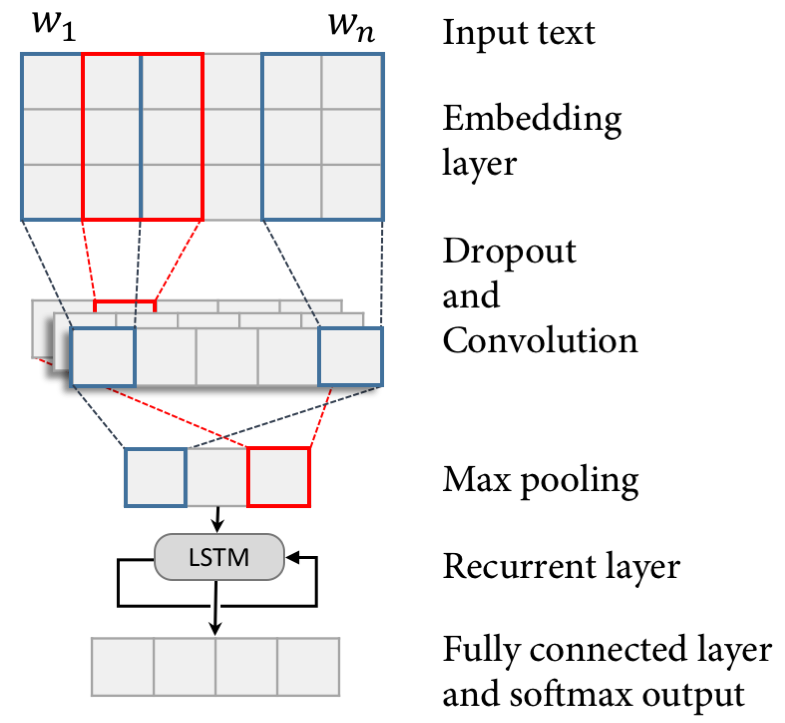}
\end{center}
    \caption{Text classification model architecture.}
    \label{fig:architecture}
\end{figure}

In our experiments, we use a standard neural network based classification approach which is similar to the Convolution Neural Network classifier of \newcite{kim2014convolutional} and the pioneering model of \newcite{Collobert:2011}.
Figure \ref{fig:architecture} depicts the architecture of the model.
The network receives the concatenated vector representations of the input words,
$\mathbf{v}_{1:n} = \mathbf{v}_1 \oplus \mathbf{v}_2 \oplus \dots \oplus \mathbf{v}_n$, 
\noindent and applies (convolves) filters $\mathrm{F}$ on windows of $h$ words,
$m_i=f(\mathrm{F}.\mathbf{v}_{i:i+h-1} + b)$,
where $b$ is a bias term and $f()$ is a non-linear function, for which we use ReLU \cite{icml2010_NairH10}.
The convolution transforms the input text to a feature map
$m=[m_1,m_2,\dots,m_{n-h+1}]$.
A max pooling operation then selects the most salient feature $\hat{m} = \mathrm{max}\{m\}$ for each filter.

In the network of \newcite{kim2014convolutional}, the pooled features are directly passed to a fully connected
softmax layer whose outputs are class probabilities.
However, we add a recurrent layer before softmax in order to enable better capturing of long-distance dependencies.
It has been shown by \newcite{XiaoCho:2016} that a recurrent layer can replace multiple layers of convolution and be beneficial, particularly when the length of input text grows.
Specifically, we use a Long Short-Term Memory \cite[LSTM]{hochreiter1997long} as our recurrent layer which was originally proposed to avoid the vanishing gradient problem and has proven its abilities in capturing distant dependencies. 
The LSTM unit computes three gate vectors (forget, input, and output) as follows:
\begin{equation}
\begin{aligned}
\mathbf{f}_t & = \sigma(\mathbf{W}_f ~ g_t + \mathbf{U}_f ~ h_{t-1} + \mathbf{b}_f),\\
\mathbf{i}_t & = \sigma(\mathbf{W}_i ~ g_t + \mathbf{U}_i ~ h_{t-1} + \mathbf{b}_i),\\
\mathbf{o}_t & = \sigma(\mathbf{W}_o ~ g_t + \mathbf{U}_o ~ h_{t-1} + \mathbf{b}_o), \\
\end{aligned}
\end{equation} 
where $\mathbf{W}$, $\mathbf{U}$, and $\mathbf{b}$ are model parameters and $g$ and $h$ are input and output sequences, respectively.
The cell state vector $\mathbf{c}_t$ is then computed as $\mathbf{c}_t = \mathbf{f}_t ~ \mathbf{c}_{t-1} + \mathbf{i}_t ~ \mathrm{tanh}(\tilde{\mathbf{c}}_t)
$
\noindent where $\tilde{c}_t = \mathbf{W}_c ~ g_t + \mathbf{U}_c ~ h_{t-1}$.
Finally, the output sequence is computed as 
$h_t = \mathbf{o}_t ~ \mathrm{tanh}(\mathbf{c}_t)$.
As for regularization, we used dropout \cite{dropout:2012} after the embedding layer.

We perform experiments with two configurations of the embedding layer: (1) \textit{Random}, initialized randomly and updated during training, and (2) \textit{Pre-trained}, initialized by pre-trained representations and updated during training.
In the following section we describe the pre-trained word and sense representation used for the initialization of the second configuration.

\subsection{Pre-trained Word and Sense Embeddings}
\label{embeddings}

One of the main advantages of neural models is that they usually represent the input words as dense vectors.
This can significantly boost a system's generalisation power and results in improved performance \cite[\textit{interalia}]{zou2013bilingual,bordes2014question,kim2014convolutional,weiss2015structured}. This feature also enables us to directly plug in pre-trained sense representations and check them in a downstream application.

In our experiments we generate a set of sense embeddings by extending DeConf, a recent technique with state-of-the-art performance on multiple semantic similarity benchmarks \cite{PilehvarCollier:2016emnlp}.
We leave the evaluation of other representations to future work.
DeConf gets a pre-trained set of word embeddings and computes sense embeddings in the same semantic space. To this end, the approach exploits the semantic network of WordNet \cite{miller1995wordnet}, using the Personalized PageRank \cite{Haveliwala:02} algorithm, and obtains a set of \textit{sense biasing words} $\mathcal{B}_s$ for a word sense $s$.
The sense representation of $s$ is then obtained using the following formula:

\begin{equation}
\label{equation_deconf}
\hat{\mathbf{v}}(s) = \frac{1}{|\mathcal{B}_s|} \sum_{i = 1}^{|\mathcal{B}_s|} \mathrm{e}^{\frac{-i}{\delta}}\mathbf{v}(w_i),    
\end{equation}

\noindent where $\delta$ is a decay parameter and $\mathbf{v}(w_i)$ is the embedding of $w_i$, i.e. the $i^{th}$ word in the sense biasing list of $s$, i.e. $\mathcal{B}_s$. 
We follow \newcite{PilehvarCollier:2016emnlp} and set $\delta=5$. 
Finally, the vector for sense $s$ is calculated as the average of $\hat{\mathbf{v}}(s)$ and the embedding of its corresponding word.

Owing to its reliance on WordNet's semantic network, DeConf is limited to generating only those word senses that are covered by this lexical resource. We propose to use Wikipedia in order to expand the vocabulary of the computed word senses. Wikipedia provides a high coverage of named entities and domain-specific terms in many languages, while at the same time also benefiting from a continuous update by collaborators. Moreover, it can easily be viewed as a sense inventory where individual articles are word senses arranged through hyperlinks and redirections.

\newcite{camacho2016nasari} proposed \textsc{Nasari}\footnote{We downloaded the salient words for Wikipedia pages (\textsc{Nasari} English lexical vectors, version 3.0) from \url{http://lcl.uniroma1.it/nasari/}}, a technique to compute the most salient words for each Wikipedia page. These salient words were computed by exploiting the structure and content of Wikipedia and proved effective in tasks such as Word Sense Disambiguation \cite{tripodi2017game,CAMACHOCOLLADOS16.629}, knowledge-base construction \cite{lieto2016resource}, domain-adapted hypernym discovery \cite{EspinosaEMNLP2016,babeldomains:2017} or object recognition \cite{young2016towards}. 
We view these lists as \textit{biasing words} for individual Wikipedia pages, and then leverage the exponential decay function (Equation \ref{equation_deconf}) to compute new sense embeddings in the same semantic space.
In order to represent both WordNet and Wikipedia sense representations in the same space, we rely on the WordNet-Wikipedia mapping provided by BabelNet\footnote{We used the Java API from \url{http://babelnet.org}} \cite{NavigliPonzetto:12aij}. For the WordNet synsets which are mapped to Wikipedia pages in BabelNet, we average the corresponding Wikipedia-based and WordNet-based sense embeddings.

\subsection{Pre-trained Supersense Embeddings}
\label{sec:supersenses}

It has been argued that WordNet sense distinctions are too fine-grained for many NLP applications \cite{Hovyetal:13}. The issue can be tackled by grouping together similar senses of the same word, either using automatic clustering techniques \cite{Navigli:06b,AgirreLopez:03,Snowetal:2007} or with the help of WordNet's lexicographer files\footnote{\url{https://wordnet.princeton.edu/man/lexnames.5WN.html}}. Various applications have been shown to improve upon moving from senses to supersenses \cite{rud-EtAl:2011:ACL-HLT2011,severyn-nicosia-moschitti:2013:Short,flekovasupersense}. In WordNet's lexicographer files there are a total of 44 sense clusters, referred to as supersenses, for categories such as \textit{event}, \textit{animal}, and \textit{quantity}. In our experiments we use these supersenses in order to reduce granularity of our WordNet and Wikipedia senses. To generate supersense embeddings, we simply average the embeddings of senses in the corresponding cluster.


\section{Evaluation}
\label{sec:evaluation}

We evaluated our model on two classification tasks: topic categorization (Section \ref{ev:topic}) and polarity detection (Section \ref{ev:polarity}). In the following section we present the common experimental setup.

\subsection{Experimental setup}
\label{setup}

\paragraph{Classification model.} 
Throughout all the experiments we used the classification model described in Section \ref{model}. 
The general architecture of the model was the same for both tasks, with slight variations in hyperparameters given the different natures of the tasks, following the values suggested by  \newcite{kim2014convolutional} and \newcite{XiaoCho:2016} for the two tasks. Hyperparameters were fixed across all configurations in the corresponding tasks.
The embedding layer was fixed to 300 dimensions, irrespective of the configuration, i.e. Random and Pre-trained.
For both tasks the evaluation was carried out by 10-fold cross-validation unless standard training-testing splits were available. The disambiguation threshold $\theta$ (cf. Section \ref{disambiguation}) was tuned on the training portion 
of the corresponding data, over seven values in [0,3] in steps of 0.5.\footnote{We observed that values higher than 3 led to very few disambiguations. While the best results were generally achieved in the [1.5,2.5] range, performance differences across threshold values were not statistically significant in most cases.}  
We used Keras \cite{chollet2015} and Theano \cite{2016arXiv160502688short} for our model implementations.


\paragraph{Semantic network.} The integration of senses was carried out as described in Section \ref{disambiguation}. For disambiguating with both WordNet and Wikipedia senses we relied on the joint semantic network of Wikipedia hyperlinks and WordNet via the mapping provided by BabelNet.\footnote{For simplicity we refer to this joint sense inventory as Wikipedia, but note that WordNet senses are also covered.}

\paragraph{Pre-trained word and sense embeddings.} Throughout all the experiments we used Word2vec \cite{Mikolovetal:2013} embeddings, trained on the Google News corpus.\footnote{\url{ https://code.google.com/archive/p/word2vec/}}
We truncated this set to its 250K most frequent words.
We also used WordNet 3.0 \cite{Fellbaum:98} and the Wikipedia dump of November 2014 to compute the sense embeddings (see Section \ref{embeddings}). As a result, we obtained a set of 757,262 sense embeddings in the same space as the pre-trained Word2vec word embeddings. We used DeConf \cite{PilehvarCollier:2016emnlp} as our pre-trained WordNet sense embeddings. All vectors had a fixed dimensionality of 300.


\paragraph{Supersenses.} In addition to WordNet senses, we experimented with supersenses (see Section \ref{sec:supersenses}) to check how reducing granularity would affect system performance. For obtaining supersenses in a given text we relied on our disambiguation pipeline and simply clustered together senses belonging to the same WordNet supersense.

\paragraph{Evaluation measures.} We report the results in terms of standard accuracy and F1 measures.\footnote{Since all models in our experiments provide full coverage, accuracy and F1 denote micro- and macro-averaged F1, respectively \cite{yang1999evaluation}.}

\subsection{Topic Categorization}
 \label{ev:topic}

The task of topic categorization consists of assigning a label (i.e. topic) to a given document from a pre-defined set of labels. 


\begin{table*}
\begin{center}
{
\setlength{\tabcolsep}{6pt}
\scalebox{0.82}{ 
\begin{tabular}{llcrrrcc}
\toprule
\bf Dataset &   \bf Domain & \bf  No. of classes & \bf No. of docs & \bf Avg. doc. size & \bf Size of vocab. & \bf Coverage & \bf Evaluation \\
\midrule
\bf BBC &  News & 5 &  2,225 & 439.5 & 35,628   & 87.4\%    & 10 cross valid.  \\
\bf Newsgroups   &   News   &   6   &   18,846  &   394.0   &  225,046 &    83.4\% &   Train-Test  \\

\bf Ohsumed &   Medical &   23  &   23,166  &   201.2   &   65,323 & 79.3\%    & Train-Test    \\


\bottomrule
\end{tabular}
}
}
\end{center}
\caption{\label{tab:statscategorization} Statistics of the topic categorization datasets.}
\end{table*}

\begin{table*}[t]
\renewcommand{\arraystretch}{0.6}
\setlength{\tabcolsep}{10.0pt}
\begin{center}
\scalebox{0.9}{

    \centering
    \begin{tabular}{l  l  l  c  c  c  c c  c }
        \toprule
      \multicolumn{1}{l}{\multirow{2}{*}{{\textbf{Initialization}}}} & \multicolumn{2}{l}{\multirow{2}{*}{{\textbf{Input type}}}}   &  \multicolumn{2}{c}{\textbf{BBC News}} & \multicolumn{2}{c}{\textbf{Newsgroups}} & \multicolumn{2}{c}{\textbf{Ohsumed}}  \\
         \cmidrule(lr){4-5}
         \cmidrule(lr){6-7}
         \cmidrule(lr){8-9}
       & \multicolumn{1}{c}{}  & & \multicolumn{1}{c}{\textbf{Acc}}  & \multicolumn{1}{c}{\textbf{F1}} & \multicolumn{1}{c}{\textbf{Acc}}  & \multicolumn{1}{c}{\textbf{F1}} & \multicolumn{1}{c}{\textbf{Acc}}  & \multicolumn{1}{c}{\textbf{F1}}  \\
        \midrule
        \multirow{8}{*}{\textbf{Random}} & \multicolumn{2}{l}{\textbf{Word}} & 93.0 & 92.8 & 87.7 & 85.6 & 30.1 & 20.7 \\
         \cmidrule{2-9}
         \cmidrule{2-9}
        & \multirow{2}{*}{\textbf{Sense}} & 
        \textbf{WordNet} & \bf 93.5 & \bf 93.3 & \bf 88.1 & \bf 86.9 & ~~27.2$^\dagger$ & 18.3 \\
         \cmidrule{3-9}
       & &  \textbf{Wikipedia} & 92.7 & 92.5 & 86.7 & 84.9 & 29.7 & \bf 20.9 \\
        \cmidrule{2-9}
       & \multirow{2}{*}{\textbf{Supersense}} & 
        \textbf{WordNet} & \bf 93.6 & \bf 93.4 & \bf ~~90.1$^*$ & \bf 89.0 & \bf ~~31.8$^*$ & \bf 22.0 \\
        \cmidrule{3-9}
       & &  \textbf{Wikipedia} & \bf ~~94.6$^*$ & \bf 94.4 & \bf 88.5 & \bf 85.8 & \bf 31.1 & \bf 21.3 \\

        \midrule
              \midrule

       \multirow{8}{*}{\textbf{Pre-trained}} & \multicolumn{2}{l}{\textbf{Word}} & 97.6 & 97.5 & 91.1 & 90.6 & 29.4 & 20.1 \\
        \cmidrule{2-9}
        \cmidrule{2-9}
        & \multirow{2}{*}{\textbf{Sense}} & 
        \textbf{WordNet} & 97.3 & 97.1 & 90.2 & 88.6 & \bf 30.2 & \bf 20.4 \\
         \cmidrule{3-9}
       & &  \textbf{Wikipedia} & 96.3 & 96.2 & ~~89.6$^\dagger$ & 88.9 & \bf 32.4 & \bf 22.3 \\
        \cmidrule{2-9}
       & \multirow{2}{*}{\textbf{Supersense}} & 
        \textbf{WordNet} & 96.8 & 96.7 & 89.6 & 88.9 & \bf 29.5 & 19.9 \\
        \cmidrule{3-9}
       & &  \textbf{Wikipedia} & 96.9 & 96.9 & 88.6 & 87.4 & \bf ~~30.6$^*$ & \bf 20.3 \\

       \bottomrule
    \end{tabular}
    }
     
     \end{center}
    \caption{Classification performance at the word, sense, and supersense levels with random and pre-trained embedding initialization. We show in bold those settings that improve the word-based model.}
    \bigskip
    \label{tab:results-cat}

\end{table*}

\subsubsection{Datasets}

For this task we used two newswire and one medical topic categorization datasets. 
Table \ref{tab:statscategorization} summarizes the statistics of each dataset.\footnote{The coverage of the datasets was computed using the 250K top words in the Google News Word2vec embeddings.}
The \textbf{BBC news} dataset\footnote{\url{http://mlg.ucd.ie/datasets/bbc.html}} \cite{greene2006practical} comprises news articles taken from BBC, divided into five topics: business, entertainment, politics, sport and tech. \textbf{Newsgroups} \cite{lang1995newsweeder} is a collection of 11,314 documents for training and 7532 for testing\footnote{We used the train-test partition available at \url{http://qwone.com/~jason/20Newsgroups/}} divided into six topics: computing, sport and motor vehicles, science, politics, religion and sales.\footnote{The dataset has 20 fine-grained categories clustered into six general topics. We used the coarse-grained labels for their clearer distinction and consistency with BBC topics.} Finally, \textbf{Ohsumed}\footnote{\url{ftp://medir.ohsu.edu/pub/ohsumed}} is a collection of medical abstracts from MEDLINE, an online medical information database, categorized according to 23 cardiovascular diseases. For our experiments we used the partition split of 10,433 documents for training and 12,733 for testing.\footnote{\url{http://disi.unitn.it/moschitti/corpora.htm}}

\subsubsection{Results}
 
Table \ref{tab:results-cat} shows the results of our classification model and its variants on the three datasets.\footnote{Symbols $^*$ and $^\dagger$ indicate the sense-based model with the smallest margin to the word-based model whose accuracy is statistically significant at 0.95 confidence level according to unpaired t-test ($^*$ for positive and $^\dagger$ for negative change).}
When the embedding layer is initialized randomly, the model integrated with word senses
consistently improves over the word-based model, particularly when the fine-granularity of the underlying sense inventory is reduced using supersenses (with statistically significant gains on the three datasets).
This highlights the fact that a simple disambiguation of the input can bring about performance gain for a state-of-the-art classification system.
Also, the better performance of supersenses suggests that the sense distinctions of WordNet are too fine-grained for the topic categorization task. However, when pre-trained representations are used to initialize the embedding layer, no improvement is observed over the word-based model.
This can be attributed to the quality of the representations, as the model utilizing them was unable to benefit from the advantage offered by sense distinctions.
Our results suggest that research in sense representation should put special emphasis on real-world evaluations on benchmarks for downstream applications, rather than on artificial tasks such as word similarity. In fact, research has previously shown that word similarity might not constitute a reliable proxy to measure the performance of word embeddings in downstream applications  \cite{tsvetkov2015evaluation,chiu2016intrinsic}.

Among the three datasets, Ohsumed proves to be the most challenging one, mainly for its larger number of classes (i.e. 23) and its domain-specific nature (i.e. medicine).
Interestingly, unlike for the other two datasets, the introduction of pre-trained word embeddings to the system results in reduced performance on Ohsumed. This suggests that general domain embeddings might not be beneficial in specialized domains, which corroborates previous findings by \newcite{yadav-EtAl:2017:EACLlong} on a different task, i.e. entity extraction. This performance drop may also be due to diachronic issues (Ohsumed dates back to the 1980s) and low coverage: the pre-trained Word2vec embeddings cover 79.3\% of the words in Ohsumed (see Table \ref{tab:statscategorization}), in contrast to the higher coverage on the newswire datasets, i.e. Newsgroups (83.4\%) and BBC (87.4\%). However, also note that the best overall performance is attained when our pre-trained Wikipedia sense embeddings are used. This highlights the effectiveness of Wikipedia in handling domain-specific entities, thanks to its broad sense inventory. 



\begin{table*}

\begin{center}
{
\setlength{\tabcolsep}{11pt}
\scalebox{0.84}{ 
\begin{tabular}{llrrrcc}
\toprule
 \multicolumn{1}{c}{\textbf{Dataset}} & \multicolumn{1}{c}{\textbf{Type}}&  \multicolumn{1}{c}{\bf No. of docs} &  \multicolumn{1}{c}{\bf Avg. doc. size}
 &\bf Vocabulary size & \bf  Coverage &\multicolumn{1}{c}{\bf Evaluation}     \\
\midrule

\bf RTC & Snippets &  438,000    & 23.4 & 128,056    &   81.3\%  &   Train-Test  \\
\bf IMDB	& Reviews   &  50,000   & 268.8 & 140,172 & 82.5\% & Train-Test  \\
\bf PL05 & Snippets  &  10,662    & 21.5 
& 19,825 & 81.3\% & 10 cross valid.     \\
\bf PL04 & Reviews   &  2,000    &  762.1 & 45,077    &   82.4\% & 10 cross valid.     \\
\bf Stanford & Phrases  &  119,783    & 10.0 & 19,400 & 81.6\% & 10 cross valid.     \\
\bottomrule

\end{tabular}
}
}
\end{center}
\caption{\label{tab:statspolarity} Statistics of the polarity detection datasets.}
\end{table*}

\begin{table*}[t]
\begin{center}
\renewcommand{\arraystretch}{0.6}
\setlength{\tabcolsep}{10.0pt}
\scalebox{0.94}{
      \begin{tabular}{l l l  c  c  c  c c}
        \toprule
      \bf Initialization & \multicolumn{2}{l}{\textbf{Input type}}   &  
      \multicolumn{1}{c}{\textbf{RTC}} &
      \multicolumn{1}{c}{\textbf{IMDB}} &
      \multicolumn{1}{c}{\textbf{PL05}} &  \multicolumn{1}{c}{\textbf{PL04}} & 
      \multicolumn{1}{c}{\textbf{Stanford}}\\
    \midrule
    \multirow{8}{*}{\textbf{Random}} & \multicolumn{2}{l}{\textbf{Word}} & 83.6  & 87.7 & 77.3 & 67.9 & 91.8 \\
        \cmidrule{2-8}
        \cmidrule{2-8}
        & \multirow{2}{*}{\textbf{Sense}} & 
        \textbf{WordNet} & 83.2 & 87.4 & 76.6 & 67.4 & 91.3 \\
         \cmidrule{3-8}
       & &  \textbf{Wikipedia} & 83.1 &\bf 88.0 & ~~75.9$^\dagger$ & 67.1 & 91.0 \\
        \cmidrule{2-8}
       & \multirow{2}{*}{\textbf{Supersense}} & 
        \textbf{WordNet} & \bf 84.4 &\bf 88.0 & 75.9 & 66.2 & ~~91.4$^\dagger$\\
        \cmidrule{3-8}
       & &  \textbf{Wikipedia} &  83.1 &\bf ~~88.4$^*$ & 75.8 & \bf ~~69.3$^*$  & 91.0  \\
       
       \midrule
        \midrule
       
        \multirow{8}{*}{\textbf{Pre-trained}} & \multicolumn{2}{l}{\textbf{Word}} & 85.5 & 88.3 & 80.2 & 72.5 & 93.1\\
         \cmidrule{2-8}
         \cmidrule{2-8}
        & \multirow{2}{*}{\textbf{Sense}} & 
        \textbf{WordNet} & 83.4 & \bf 88.3 &  79.2 & ~~69.7$^\dagger$ & 92.6\\
         \cmidrule{3-8}
       &  & \textbf{Wikipedia} & 83.8 & ~~87.0$^\dagger$ & 79.2 & \bf 73.1 & 92.3\\
        \cmidrule{2-8}
       & \multirow{2}{*}{\textbf{Supersense}} & 
        \textbf{WordNet} & 85.2 &\bf 88.8 & 79.5 & \bf 73.8 & ~~92.7$^\dagger$ \\
        \cmidrule{3-8}
       & &  \textbf{Wikipedia} & 84.2 & 87.9 & ~~78.3$^\dagger$ & \bf 72.6  & 92.2\\

       \bottomrule
    \end{tabular}
    }
    \end{center}
    \caption[Caption title in LOF]{Accuracy performance on five polarity detection datasets. Given that polarity datasets are balanced\footnotemark,  we do not report F1 which would have been identical to accuracy.}
    \bigskip
    \label{tab:results-random-pol}
\end{table*}

\subsection{Polarity Detection}
 \label{ev:polarity}

  Polarity detection is the most popular evaluation framework for sentiment analysis  \cite{dong2015statistical}. 
  The task is essentially a binary classification which determines if the sentiment of a given sentence or document is negative or positive.

 \subsubsection{Datasets}

For the polarity detection task we used five standard evaluation datasets. Table \ref{tab:statscategorization} summarizes statistics. \textbf{PL04} \cite{Pang+Lee:04a} is a polarity detection dataset composed of full movie reviews. \textbf{PL05}\footnote{Both PL04 and PL05 were downloaded from \url{http://www.cs.cornell.edu/people/pabo/movie-review-data/}} \cite{Pang+Lee:05a}, instead, is composed of short snippets from movie reviews. \textbf{RTC} contains critic reviews from Rotten Tomatoes\footnote{\url{http://www.rottentomatoes.com}}, divided into 436,000 training and 2,000 test instances. \textbf{IMDB} \cite{maas-EtAl:2011:ACL-HLT2011} includes 50,000 movie reviews, split evenly between training and test. 
Finally, we used the \textbf{Stanford} Sentiment dataset \cite{SocherEtAl2013:RNTN}, which associates each review with a value that denotes its sentiment. To be consistent with the binary classification of the other datasets, we removed the neutral phrases according to the dataset's scale (between 0.4 and 0.6) and considered the reviews whose values were below 0.4 as negative and above 0.6 as positive. This resulted in a binary polarity dataset of 119,783 phrases. Unlike the previous four datasets, this dataset does not contain an even distribution of positive and negative labels.

\subsubsection{Results}

Table \ref{tab:results-random-pol} lists accuracy performance of our classification model and all its variants on five polarity detection datasets. 
Results are generally better than those of \newcite{kim2014convolutional}, showing that the addition of the recurrent layer to the model (cf. Section \ref{model}) was beneficial.
However, interestingly, no consistent performance gain is observed in the polarity detection task, when the model is provided with disambiguated input, particularly for datasets with relatively short reviews.
We attribute this to the nature of the task.
Firstly, given that words rarely happen to be ambiguous with respect to their sentiment,
the semantic sense distinctions provided by the disambiguation stage do not assist the classifier in better decision making, and instead introduce data sparsity. 
Secondly, since the datasets mostly contain short texts, e.g., sentences or snippets, the disambiguation algorithm does not have sufficient context to make high-confidence judgements, resulting in fewer disambiguations or less reliable ones.
In the following section we perform a more in-depth analysis of the impact of document size on the performance of our sense-based models. 

\footnotetext{Stanford is the only unbalanced dataset, but F1 results were almost identical to accuracy.}

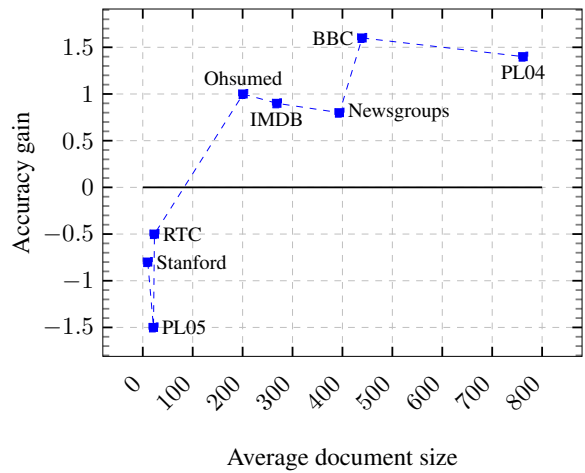
\begin{figure}
\centering
\begin{tikzpicture}[scale=0.85]
\begin{axis}[
    width = 9cm,
    height = 7cm,
    xlabel={Average document size},
    xlabel style={yshift=-2em},
    ylabel={Accuracy gain},
    grid style=dashed,
    ymajorgrids=true,
    xmajorgrids=true,
    scaled y ticks=false,
    minor y tick num=4,
    major tick style={draw=black, thick},
    minor tick style={thick},
    max space between ticks = 25,
    x tick label style={rotate=45,anchor=east, xshift=-0.5em, yshift=-0.5em},
    legend style={legend pos=south east, font=\small, legend cell align=left}
]

\addplot[color=blue, mark = square*, dashed]
    coordinates {
    (10.0, -0.8)
    (21.5, -1.5)
    (23.4, -0.5)
    (201.2, 1)
    (268.8, 0.9)
    (394, 0.8)
    (439.5, 1.6)
    (762.1, 1.4)
    };
    
\node[black,right] at (axis cs:10.0,-0.8){\small{Stanford}};
\node[black,right] at (axis cs:21.5,-1.5){\small{PL05}};
\node[black,right] at (axis cs:23.4, -0.5){\small{RTC}};
\node[black,above] at (axis cs:201.2, 1){\small{Ohsumed}};
\node[black,below] at (axis cs:268.8, 0.9){\small{IMDB}};
\node[black,right] at (axis 
cs:394, 0.8){\small{Newsgroups}};
\node[black,left] at (axis cs:439.5, 1.6){\small{BBC}};
\node[black,below] at (axis cs:762.1, 1.4){\small{PL04}};

 \addplot[mark=none, black, samples=2, thick] coordinates {
    (0, 0)
    (800, 0)};
    
\end{axis}
\end{tikzpicture}
\caption{Relation between average document size and performance improvement using Wikipedia supersenses with random initialization.}
\label{fig:analysis2}
\end{figure}

\subsection{Analysis}

\textbf{Document size.} A detailed analysis revealed a relation between document size (the number of tokens) and performance gain of our sense-level model.
We show in Figure \ref{fig:analysis2} how these two vary for our most consistent configuration, i.e. Wikipedia supersenses, with random initialization.
Interestingly, as a general trend, the performance gain increases with average document size, irrespective of the classification task.
We attribute this to two main factors:

\begin{enumerate}
    
\item \textbf{Sparsity}: 
Splitting a word into multiple word senses can have the negative side effect that the corresponding training data for that word is distributed among multiple independent senses. This reduces the training instances per word sense, which might affect the classifier's performance, particularly when senses are semantically related (in comparison to fine-grained senses, supersenses address this issue to some extent). 


\item \textbf{Disambiguation quality}: 
As also mentioned previously, our disambiguation algorithm requires the input text to be sufficiently large so as to create a graph with an adequate number of coherent connections to function effectively. In fact, for topic categorization, in which the documents are relatively long, our algorithm manages to disambiguate a larger proportion of words in documents with high confidence. The lower performance of graph-based disambiguation algorithms on short texts is a known issue \cite{Moroetal:14tacl,raganato-camachocollados-navigli:2017:EACLlong}, the tackling of which remains an area of exploration.


\end{enumerate}

\paragraph{Senses granularity.} Our results showed that reducing fine-granularity of sense distinctions can be beneficial to both tasks, irrespective of the underlying sense inventory, i.e. WordNet or Wikipedia, which corroborates previous findings \cite{Hovyetal:13,flekovasupersense}. 
This suggests that text classification does not require fine-grained semantic distinctions.
In this work we used a simple technique based on WordNet's lexicographer files for coarsening senses in this sense inventory as well as in Wikipedia.
We leave the exploration of this promising area as well as the evaluation of other granularity reduction techniques for WordNet \cite{Snowetal:2007,bhagwani-satapathy-karnick:2013:TextGraphs} and Wikipedia \cite{Dandalaetal:2013} sense inventories to future work.

\section{Related Work}

The past few years have witnessed a growing research interest in semantic representation, mainly as a consequence of the word embedding tsunami \cite{Mikolovetal:2013,pennington2014glove}.
Soon after their introduction, word embeddings were integrated into different NLP applications, thanks to the migration of the field to deep learning and the fact that most deep learning models view words as dense vectors. 
The waves of the word embedding tsunami have also lapped on the shores of sense representation.
Several techniques have been proposed that either extend word embedding models to cluster contexts and induce senses, usually referred to as unsupervised sense representations \cite{schutze1998automatic,ReisingerMooney:2010,Huangetal:2012,Neelakantanetal:2014,guo2014learning,tian2014probabilistic,vsuster2016bilingual,ettinger2016retrofitting,qiu-tu-yu:2016:EMNLP2016} or exploit external sense inventories and lexical resources for generating sense representations for individual meanings of words \cite{chenunified:2014,johansson2015embedding,jauhar2015ontologically,iacobacci:2015,RotheSchutze:2015,camacho2016nasari,mancini2016sw2v,PilehvarCollier:2016emnlp}.

However, the integration of sense representations into deep learning models has not been so straightforward, and research in this field has often opted for alternative evaluation benchmarks such as WSD, or artificial tasks, such as word similarity. 
Consequently, the problem of integrating sense representations into downstream NLP applications has remained understudied, despite the potential benefits it can have.
\newcite{LiJurafsky:2015} proposed a ``multi-sense embedding" pipeline to check the benefit that can be gained by replacing word embeddings with sense embeddings in multiple tasks.
With the help of two simple disambiguation algorithms,  unsupervised sense embeddings were integrated into various downstream applications, with varying degrees of success. 
Given the interdependency of sense representation and disambiguation in this model, it is very difficult to introduce alternative algorithms into its pipeline, either to benefit from the state of the art, or to carry out an evaluation.  
Instead, our pipeline provides the advantage of being modular: thanks to its use of disambiguation in the pre-processing stage and use of sense representations that are linked to external sense inventories, different WSD techniques and sense representations can be easily plugged in and checked.
Along the same lines, \newcite{flekovasupersense} proposed a technique for learning supersense representations, using automatically-annotated corpora.
Coupled with a supersense tagger, the representations were fed into a neural network classifier as additional features to the word-based input.
Through a set of experiments, \newcite{flekovasupersense} showed that the supersense enrichment can be beneficial to a range of binary classification tasks.
Our proposal is different in that it focuses directly on the benefits that can be gained by semantifying the input, i.e. reducing lexical ambiguity in the input text, rather than assisting the model with additional sources of knowledge.

\section{Conclusion and Future Work}
\label{sec:length}

We proposed a pipeline for the integration of sense level knowledge into a state-of-the-art text classifier.
We showed that a simple disambiguation of the input can lead to consistent performance gain, particularly for longer documents and when the granularity of the underlying sense inventory is reduced.
Our pipeline is modular and can be used as an \textit{in vivo} evaluation framework for WSD and sense representation techniques.
We release our code and data to reproduce our experiments (including pre-trained sense and supersense embeddings) at 
\url{https://github.com/pilehvar/sensecnn} to allow further checking of the choice of hyperparameters and to allow further analysis and comparison.
We hope that our work will foster future research on the integration of sense-level knowledge into downstream applications. As future work, we plan to investigate the extension of the approach to other languages and applications. Also, given the promising results observed for supersenses, we will investigate task-specific coarsening of sense inventories, particularly Wikipedia, or the use of SentiWordNet \cite{baccianella2010sentiwordnet}, which could be more suitable for polarity detection.







\input{acknowledgements}

\bibliography{acl2017}
\bibliographystyle{acl_natbib}

\appendix

\end{document}

%% file: acknowledgements.tex

\section*{Acknowledgments}

The authors gratefully acknowledge the support of the MRC grant No. MR/M025160/1 for PheneBank and ERC Consolidator Grant MOUSSE No. 726487. Jose Camacho-Collados is supported by a Google Doctoral Fellowship in Natural Language Processing. Nigel Collier is supported by EPSRC Grant No. EP/M005089/1.
We thank Jim McManus for his suggestions on the manuscript and the anonymous reviewers for their helpful comments.

%% file: acl2017.bbl
\begin{thebibliography}{}
\expandafter\ifx\csname natexlab\endcsname\relax\def\natexlab#1{#1}\fi

\bibitem[{Agirre et~al.(2014)Agirre, de~Lacalle, and Soroa}]{agirre2014random}
Eneko Agirre, Oier~Lopez de~Lacalle, and Aitor Soroa. 2014.
\newblock Random walks for knowledge-based word sense disambiguation.
\newblock {\em Computational Linguistics\/} 40(1):57--84.

\bibitem[{Agirre and Lopez(2003)}]{AgirreLopez:03}
Eneko Agirre and Oier Lopez. 2003.
\newblock Clustering {W}ord{N}et word senses.
\newblock In {\em Proceedings of Recent Advances in Natural Language
  Processing\/}. Borovets, Bulgaria, pages 121--130.

\bibitem[{Baccianella et~al.(2010)Baccianella, Esuli, and
  Sebastiani}]{baccianella2010sentiwordnet}
Stefano Baccianella, Andrea Esuli, and Fabrizio Sebastiani. 2010.
\newblock Sentiwordnet 3.0: An enhanced lexical resource for sentiment analysis
  and opinion mining.
\newblock In {\em LREC\/}. volume~10, pages 2200--2204.

\bibitem[{Ballesteros et~al.(2015)Ballesteros, Dyer, and
  Smith}]{ballesterosimproved}
Miguel Ballesteros, Chris Dyer, and Noah~A Smith. 2015.
\newblock Improved transition-based parsing by modeling characters instead of
  words with lstms.
\newblock In {\em Proceedings of EMNLP\/}.

\bibitem[{Bhagwani et~al.(2013)Bhagwani, Satapathy, and
  Karnick}]{bhagwani-satapathy-karnick:2013:TextGraphs}
Sumit Bhagwani, Shrutiranjan Satapathy, and Harish Karnick. 2013.
\newblock Merging word senses.
\newblock In {\em Proceedings of {TextGraphs}-8 Graph-based Methods for Natural
  Language Processing\/}. Seattle, Washington, USA, pages 11--19.

\bibitem[{Bordes et~al.(2014)Bordes, Chopra, and Weston}]{bordes2014question}
Antoine Bordes, Sumit Chopra, and Jason Weston. 2014.
\newblock Question answering with subgraph embeddings.
\newblock In {\em EMNLP\/}.

\bibitem[{Camacho-Collados et~al.(2016{\natexlab{a}})Camacho-Collados, Bovi,
  Raganato, and Navigli}]{CAMACHOCOLLADOS16.629}
Jos\'{e} Camacho-Collados, Claudio~Delli Bovi, Alessandro Raganato, and Roberto
  Navigli. 2016{\natexlab{a}}.
\newblock {A Large-Scale Multilingual Disambiguation of Glosses}.
\newblock In {\em Proceedings of LREC\/}. Portoroz, Slovenia, pages 1701--1708.

\bibitem[{Camacho-Collados and Navigli(2017)}]{babeldomains:2017}
Jose Camacho-Collados and Roberto Navigli. 2017.
\newblock {BabelDomains: Large-Scale Domain Labeling of Lexical Resources}.
\newblock In {\em Proceedings of EACL (2)\/}. Valencia, Spain.

\bibitem[{Camacho-Collados et~al.(2016{\natexlab{b}})Camacho-Collados,
  Pilehvar, and Navigli}]{camacho2016nasari}
Jos{\'e} Camacho-Collados, Mohammad~Taher Pilehvar, and Roberto Navigli.
  2016{\natexlab{b}}.
\newblock Nasari: Integrating explicit knowledge and corpus statistics for a
  multilingual representation of concepts and entities.
\newblock {\em Artificial Intelligence\/} 240:36--64.

\bibitem[{Chen et~al.(2014)Chen, Liu, and Sun}]{chenunified:2014}
Xinxiong Chen, Zhiyuan Liu, and Maosong Sun. 2014.
\newblock A unified model for word sense representation and disambiguation.
\newblock In {\em Proceedings of EMNLP\/}. Doha, Qatar, pages 1025--1035.

\bibitem[{Cheng and Roth(2013)}]{cheng2013relational}
Xiao Cheng and Dan Roth. 2013.
\newblock Relational inference for wikification.
\newblock In {\em Proceedings of EMNLP\/}. Seattle, Washington, pages
  1787--1796.

\bibitem[{Chiu et~al.(2016)Chiu, Korhonen, and Pyysalo}]{chiu2016intrinsic}
Billy Chiu, Anna Korhonen, and Sampo Pyysalo. 2016.
\newblock Intrinsic evaluation of word vectors fails to predict extrinsic
  performance.
\newblock In {\em Proceedings of the Workshop on Evaluating Vector Space
  Representations for NLP, ACL\/}.

\bibitem[{Chollet(2015)}]{chollet2015}
Fran\c{c}ois Chollet. 2015.
\newblock Keras.
\newblock \url{https://github.com/fchollet/keras}.

\bibitem[{Collobert et~al.(2011)Collobert, Weston, Bottou, Karlen, Kavukcuoglu,
  and Kuksa}]{Collobert:2011}
Ronan Collobert, Jason Weston, L{\'e}on Bottou, Michael Karlen, Koray
  Kavukcuoglu, and Pavel Kuksa. 2011.
\newblock Natural language processing (almost) from scratch.
\newblock {\em J. Mach. Learn. Res.\/} 12:2493--2537.

\bibitem[{Dandala et~al.(2013)Dandala, Hokamp, Mihalcea, and
  Bunescu}]{Dandalaetal:2013}
Bharath Dandala, Chris Hokamp, Rada Mihalcea, and Razvan~C. Bunescu. 2013.
\newblock Sense clustering using {W}ikipedia.
\newblock In {\em Proceedings of Recent Advances in Natural Language
  Processing\/}. Hissar, Bulgaria, pages 164--171.

\bibitem[{Dong et~al.(2015)Dong, Wei, Liu, Zhou, and Xu}]{dong2015statistical}
Li~Dong, Furu Wei, Shujie Liu, Ming Zhou, and Ke~Xu. 2015.
\newblock A statistical parsing framework for sentiment classification.
\newblock {\em Computational Linguistics\/} 41(2):293--336.

\bibitem[{Espinosa-Anke et~al.(2016)Espinosa-Anke, Camacho-Collados,
  Delli~Bovi, and Saggion}]{EspinosaEMNLP2016}
Luis Espinosa-Anke, Jose Camacho-Collados, Claudio Delli~Bovi, and Horacio
  Saggion. 2016.
\newblock Supervised distributional hypernym discovery via domain adaptation.
\newblock In {\em Proceedings of EMNLP\/}. pages 424--435.

\bibitem[{Ettinger et~al.(2016)Ettinger, Resnik, and
  Carpuat}]{ettinger2016retrofitting}
Allyson Ettinger, Philip Resnik, and Marine Carpuat. 2016.
\newblock Retrofitting sense-specific word vectors using parallel text.
\newblock In {\em Proceedings of NAACL-HLT\/}. San Diego, California, pages
  1378--1383.

\bibitem[{Fellbaum(1998)}]{Fellbaum:98}
Christiane Fellbaum, editor. 1998.
\newblock {\em {W}ord{N}et: An Electronic Database\/}.
\newblock MIT Press, Cambridge, MA.

\bibitem[{Flekova and Gurevych(2016)}]{flekovasupersense}
Lucie Flekova and Iryna Gurevych. 2016.
\newblock Supersense embeddings: A unified model for supersense interpretation,
  prediction, and utilization.
\newblock In {\em Proceedings of ACL\/}.

\bibitem[{Greene and Cunningham(2006)}]{greene2006practical}
Derek Greene and P{\'a}draig Cunningham. 2006.
\newblock Practical solutions to the problem of diagonal dominance in kernel
  document clustering.
\newblock In {\em Proceedings of the 23rd International conference on Machine
  learning\/}. ACM, pages 377--384.

\bibitem[{Guo et~al.(2014)Guo, Che, Wang, and Liu}]{guo2014learning}
Jiang Guo, Wanxiang Che, Haifeng Wang, and Ting Liu. 2014.
\newblock Learning sense-specific word embeddings by exploiting bilingual
  resources.
\newblock In {\em COLING\/}. pages 497--507.

\bibitem[{Haveliwala(2002)}]{Haveliwala:02}
Taher~H. Haveliwala. 2002.
\newblock Topic-sensitive {PageRank}.
\newblock In {\em Proceedings of the 11th International Conference on World
  Wide Web\/}. Hawaii, USA, pages 517--526.

\bibitem[{Hinton et~al.(2012)Hinton, Srivastava, Krizhevsky, Sutskever, and
  Salakhutdinov}]{dropout:2012}
Geoffrey~E. Hinton, Nitish Srivastava, Alex Krizhevsky, Ilya Sutskever, and
  Ruslan Salakhutdinov. 2012.
\newblock Improving neural networks by preventing co-adaptation of feature
  detectors.
\newblock {\em CoRR\/} abs/1207.0580.

\bibitem[{Hochreiter and Schmidhuber(1997)}]{hochreiter1997long}
Sepp Hochreiter and J{\"u}rgen Schmidhuber. 1997.
\newblock Long short-term memory.
\newblock {\em Neural computation\/} 9(8):1735--1780.

\bibitem[{Hoffart et~al.(2012)Hoffart, Seufert, Nguyen, Theobald, and
  Weikum}]{hoffart2012kore}
Johannes Hoffart, Stephan Seufert, Dat~Ba Nguyen, Martin Theobald, and Gerhard
  Weikum. 2012.
\newblock Kore: keyphrase overlap relatedness for entity disambiguation.
\newblock In {\em Proceedings of CIKM\/}. pages 545--554.

\bibitem[{Hovy et~al.(2013)Hovy, Navigli, and Ponzetto}]{Hovyetal:13}
Eduard~H. Hovy, Roberto Navigli, and Simone~Paolo Ponzetto. 2013.
\newblock Collaboratively built semi-structured content and {A}rtificial
  {I}ntelligence: {T}he story so far.
\newblock {\em Artificial Intelligence\/} 194:2--27.

\bibitem[{Huang et~al.(2012)Huang, Socher, Manning, and Ng}]{Huangetal:2012}
Eric~H. Huang, Richard Socher, Christopher~D. Manning, and Andrew~Y. Ng. 2012.
\newblock Improving word representations via global context and multiple word
  prototypes.
\newblock In {\em Proceedings of ACL\/}. Jeju Island, Korea, pages 873--882.

\bibitem[{Iacobacci et~al.(2015)Iacobacci, Pilehvar, and
  Navigli}]{iacobacci:2015}
Ignacio Iacobacci, Mohammad~Taher Pilehvar, and Roberto Navigli. 2015.
\newblock Sensembed: Learning sense embeddings for word and relational
  similarity.
\newblock In {\em Proceedings of ACL\/}. Beijing, China, pages 95--105.

\bibitem[{Jauhar et~al.(2015)Jauhar, Dyer, and Hovy}]{jauhar2015ontologically}
Sujay~Kumar Jauhar, Chris Dyer, and Eduard Hovy. 2015.
\newblock Ontologically grounded multi-sense representation learning for
  semantic vector space models.
\newblock In {\em Proceedings of NAACL\/}. Denver, Colorado, pages 683--693.

\bibitem[{Johansson and Pina(2015)}]{johansson2015embedding}
Richard Johansson and Luis~Nieto Pina. 2015.
\newblock Embedding a semantic network in a word space.
\newblock In {\em Proceedings of NAACL\/}. Denver, Colorado, pages 1428--1433.

\bibitem[{Johnson and Zhang(2015)}]{johnson2015effective}
Rie Johnson and Tong Zhang. 2015.
\newblock Effective use of word order for text categorization with
  convolutional neural networks.
\newblock In {\em Proceedings of NAACL\/}. Denver, Colorado, pages 103--112.

\bibitem[{Kalchbrenner et~al.(2014)Kalchbrenner, Grefenstette, and
  Blunsom}]{kalchbrenner2014convolutional}
Nal Kalchbrenner, Edward Grefenstette, and Phil Blunsom. 2014.
\newblock A convolutional neural network for modelling sentences.
\newblock In {\em Proceedings of ACL\/}. Baltimore, USA, pages 655--665.

\bibitem[{Kim(2014)}]{kim2014convolutional}
Yoon Kim. 2014.
\newblock Convolutional neural networks for sentence classification.
\newblock In {\em Proceedings of EMNLP\/}. Doha, Qatar, pages 1746--1751.

\bibitem[{Kim et~al.(2016)Kim, Jernite, Sontag, and Rush}]{kim2015character}
Yoon Kim, Yacine Jernite, David Sontag, and Alexander~M Rush. 2016.
\newblock Character-aware neural language models.
\newblock In {\em Proceedings of AAAI\/}. Phoenix, Arizona, pages 2741--2749.

\bibitem[{Lang(1995)}]{lang1995newsweeder}
Ken Lang. 1995.
\newblock Newsweeder: Learning to filter netnews.
\newblock In {\em Proceedings of the 12th International Conference on Machine
  Learning\/}. Tahoe City, California, pages 331--339.

\bibitem[{Li and Jurafsky(2015)}]{LiJurafsky:2015}
Jiwei Li and Dan Jurafsky. 2015.
\newblock Do multi-sense embeddings improve natural language understanding?
\newblock In {\em Proceedings of EMNLP\/}. Lisbon, Portugal, pages 683--693.

\bibitem[{Lieto et~al.(2016)Lieto, Mensa, and Radicioni}]{lieto2016resource}
Antonio Lieto, Enrico Mensa, and Daniele~P Radicioni. 2016.
\newblock A resource-driven approach for anchoring linguistic resources to
  conceptual spaces.
\newblock In {\em AI* IA 2016 Advances in Artificial Intelligence\/}, Springer,
  pages 435--449.

\bibitem[{Limsopatham and Collier(2016)}]{limsopatham-collier:2016}
Nut Limsopatham and Nigel Collier. 2016.
\newblock Normalising medical concepts in social media texts by learning
  semantic representation.
\newblock In {\em Proceedings of ACL\/}. Berlin, Germany, pages 1014--1023.

\bibitem[{Ling et~al.(2015)Ling, Singh, and Weld}]{ling2015design}
Xiao Ling, Sameer Singh, and Daniel~S Weld. 2015.
\newblock Design challenges for entity linking.
\newblock {\em Transactions of the Association for Computational Linguistics\/}
  3:315--328.

\bibitem[{Maas et~al.(2011)Maas, Daly, Pham, Huang, Ng, and
  Potts}]{maas-EtAl:2011:ACL-HLT2011}
Andrew~L. Maas, Raymond~E. Daly, Peter~T. Pham, Dan Huang, Andrew~Y. Ng, and
  Christopher Potts. 2011.
\newblock Learning word vectors for sentiment analysis.
\newblock In {\em Proceedings of ACL-HLT\/}. Portland, Oregon, USA, pages
  142--150.

\bibitem[{Mancini et~al.(2016)Mancini, Camacho{-}Collados, Iacobacci, and
  Navigli}]{mancini2016sw2v}
Massimiliano Mancini, Jose Camacho{-}Collados, Ignacio Iacobacci, and Roberto
  Navigli. 2016.
\newblock \href{http://arxiv.org/abs/1612.02703}{Embedding words and senses
  together via joint knowledge-enhanced training}.
\newblock {\em CoRR\/} abs/1612.02703.
\newblock
  \href{http://arxiv.org/abs/1612.02703}{http://arxiv.org/abs/1612.02703}.

\bibitem[{Manning et~al.(2014)Manning, Surdeanu, Bauer, Finkel, Bethard, and
  McClosky}]{manning-EtAl:2014:P14-5}
Christopher~D. Manning, Mihai Surdeanu, John Bauer, Jenny Finkel, Steven~J.
  Bethard, and David McClosky. 2014.
\newblock The {Stanford} {CoreNLP} natural language processing toolkit.
\newblock In {\em Association for Computational Linguistics (ACL) System
  Demonstrations\/}. pages 55--60.

\bibitem[{Melamud et~al.(2016)Melamud, Goldberger, and
  Dagan}]{melamud2016context2vec}
Oren Melamud, Jacob Goldberger, and Ido Dagan. 2016.
\newblock context2vec: Learning generic context embedding with bidirectional
  lstm.
\newblock In {\em Proceedings of The 20th SIGNLL Conference on Computational
  Natural Language Learning\/}. Berlin, Germany, pages 51--61.

\bibitem[{Mikolov et~al.(2013)Mikolov, Chen, Corrado, and
  Dean}]{Mikolovetal:2013}
Tomas Mikolov, Kai Chen, Greg Corrado, and Jeffrey Dean. 2013.
\newblock Efficient estimation of word representations in vector space.
\newblock {\em CoRR\/} abs/1301.3781.

\bibitem[{Miller(1995)}]{miller1995wordnet}
George~A Miller. 1995.
\newblock {WordNet}: a lexical database for english.
\newblock {\em Communications of the ACM\/} 38(11):39--41.

\bibitem[{Moro et~al.(2014)Moro, Raganato, and Navigli}]{Moroetal:14tacl}
Andrea Moro, Alessandro Raganato, and Roberto Navigli. 2014.
\newblock {Entity Linking meets Word Sense Disambiguation: a Unified Approach}.
\newblock {\em Transactions of the Association for Computational Linguistics
  (TACL)\/} 2:231--244.

\bibitem[{Nair and Hinton(2010)}]{icml2010_NairH10}
Vinod Nair and Geoffrey~E. Hinton. 2010.
\newblock Rectified linear units improve restricted boltzmann machines.
\newblock In {\em Proceedings of the 27th International Conference on Machine
  Learning\/}. pages 807--814.

\bibitem[{Navigli(2006)}]{Navigli:06b}
Roberto Navigli. 2006.
\newblock Meaningful clustering of senses helps boost {Word Sense
  Disambiguation} performance.
\newblock In {\em Proceedings of COLING-ACL\/}. Sydney, Australia, pages
  105--112.

\bibitem[{Navigli and Ponzetto(2012)}]{NavigliPonzetto:12aij}
Roberto Navigli and Simone~Paolo Ponzetto. 2012.
\newblock {B}abel{N}et: {T}he automatic construction, evaluation and
  application of a wide-coverage multilingual semantic network.
\newblock {\em Artificial Intelligence\/} 193:217--250.

\bibitem[{Neelakantan et~al.(2014)Neelakantan, Shankar, Passos, and
  McCallum}]{Neelakantanetal:2014}
Arvind Neelakantan, Jeevan Shankar, Alexandre Passos, and Andrew McCallum.
  2014.
\newblock Efficient non-parametric estimation of multiple embeddings per word
  in vector space.
\newblock In {\em Proceedings of EMNLP\/}. Doha, Qatar, pages 1059--1069.

\bibitem[{Pang and Lee(2004)}]{Pang+Lee:04a}
Bo~Pang and Lillian Lee. 2004.
\newblock A sentimental education: Sentiment analysis using subjectivity
  summarization based on minimum cuts.
\newblock In {\em Proceedings of ACL\/}. Barcelona, Spain, pages 51--61.

\bibitem[{Pang and Lee(2005)}]{Pang+Lee:05a}
Bo~Pang and Lillian Lee. 2005.
\newblock Seeing stars: Exploiting class relationships for sentiment
  categorization with respect to rating scales.
\newblock In {\em Proceedings of ACL\/}. Ann Arbor, Michigan, pages 115--124.

\bibitem[{Pennington et~al.(2014)Pennington, Socher, and
  Manning}]{pennington2014glove}
Jeffrey Pennington, Richard Socher, and Christopher~D Manning. 2014.
\newblock Glo{V}e: Global vectors for word representation.
\newblock In {\em Proceedings of EMNLP\/}. pages 1532--1543.

\bibitem[{Pilehvar and Collier(2016)}]{PilehvarCollier:2016emnlp}
Mohammad~Taher Pilehvar and Nigel Collier. 2016.
\newblock De-conflated semantic representations.
\newblock In {\em Proceedings of EMNLP\/}. Austin, TX, pages 1680--1690.

\bibitem[{Pilehvar and Navigli(2014)}]{PilehvarNavigli:2014b}
Mohammad~Taher Pilehvar and Roberto Navigli. 2014.
\newblock A large-scale pseudoword-based evaluation framework for
  state-of-the-art {W}ord {S}ense {D}isambiguation.
\newblock {\em Computational Linguistics\/} 40(4).

\bibitem[{Qiu et~al.(2016)Qiu, Tu, and Yu}]{qiu-tu-yu:2016:EMNLP2016}
Lin Qiu, Kewei Tu, and Yong Yu. 2016.
\newblock Context-dependent sense embedding.
\newblock In {\em Proceedings of EMNLP\/}. Austin, Texas, pages 183--191.

\bibitem[{Raganato et~al.(2017)Raganato, Camacho-Collados, and
  Navigli}]{raganato-camachocollados-navigli:2017:EACLlong}
Alessandro Raganato, Jose Camacho-Collados, and Roberto Navigli. 2017.
\newblock Word sense disambiguation: A unified evaluation framework and
  empirical comparison.
\newblock In {\em Proceedings of EACL\/}. Valencia, Spain, pages 99--110.

\bibitem[{Reisinger and Mooney(2010)}]{ReisingerMooney:2010}
Joseph Reisinger and Raymond~J. Mooney. 2010.
\newblock Multi-prototype vector-space models of word meaning.
\newblock In {\em Proceedings of ACL\/}. pages 109--117.

\bibitem[{Rothe and Sch\"{u}tze(2015)}]{RotheSchutze:2015}
Sascha Rothe and Hinrich Sch\"{u}tze. 2015.
\newblock Autoextend: Extending word embeddings to embeddings for synsets and
  lexemes.
\newblock In {\em Proceedings of ACL\/}. Beijing, China, pages 1793--1803.

\bibitem[{R\"{u}d et~al.(2011)R\"{u}d, Ciaramita, M\"{u}ller, and
  Sch\"{u}tze}]{rud-EtAl:2011:ACL-HLT2011}
Stefan R\"{u}d, Massimiliano Ciaramita, Jens M\"{u}ller, and Hinrich
  Sch\"{u}tze. 2011.
\newblock Piggyback: Using search engines for robust cross-domain named entity
  recognition.
\newblock In {\em Proceedings of ACL-HLT\/}. Portland, Oregon, USA, pages
  965--975.

\bibitem[{Sag et~al.(2002)Sag, Baldwin, Bond, Copestake, and
  Flickinger}]{sag2002multiword}
Ivan~A Sag, Timothy Baldwin, Francis Bond, Ann Copestake, and Dan Flickinger.
  2002.
\newblock Multiword expressions: A pain in the neck for nlp.
\newblock In {\em International Conference on Intelligent Text Processing and
  Computational Linguistics\/}. Mexico City, Mexico, pages 1--15.

\bibitem[{Salehi et~al.(2015)Salehi, Cook, and Baldwin}]{salehi2015word}
Bahar Salehi, Paul Cook, and Timothy Baldwin. 2015.
\newblock A word embedding approach to predicting the compositionality of
  multiword expressions.
\newblock In {\em NAACL-HTL\/}. Denver, Colorado, pages 977--983.

\bibitem[{Sch{\"u}tze(1998)}]{schutze1998automatic}
Hinrich Sch{\"u}tze. 1998.
\newblock Automatic word sense discrimination.
\newblock {\em Computational linguistics\/} 24(1):97--123.

\bibitem[{Severyn et~al.(2013)Severyn, Nicosia, and
  Moschitti}]{severyn-nicosia-moschitti:2013:Short}
Aliaksei Severyn, Massimo Nicosia, and Alessandro Moschitti. 2013.
\newblock Learning semantic textual similarity with structural representations.
\newblock In {\em Proceedings of ACL (2)\/}. Sofia, Bulgaria, pages 714--718.

\bibitem[{Snow et~al.(2007)Snow, Prakash, Jurafsky, and Ng}]{Snowetal:2007}
Rion Snow, Sushant Prakash, Daniel Jurafsky, and Andrew~Y. Ng. 2007.
\newblock Learning to merge word senses.
\newblock In {\em Proceedings of EMNLP\/}. Prague, Czech Republic, pages
  1005--1014.

\bibitem[{Socher et~al.(2013)Socher, Perelygin, Wu, Chuang, Manning, Ng, and
  Potts}]{SocherEtAl2013:RNTN}
Richard Socher, Alex Perelygin, Jean Wu, Jason Chuang, Christopher Manning,
  Andrew Ng, and Christopher Potts. 2013.
\newblock Parsing with compositional vector grammars.
\newblock In {\em Proceedings of EMNLP\/}. Sofia, Bulgaria, pages 455--465.

\bibitem[{{\v{S}}uster et~al.(2016){\v{S}}uster, Titov, and van
  Noord}]{vsuster2016bilingual}
Simon {\v{S}}uster, Ivan Titov, and Gertjan van Noord. 2016.
\newblock Bilingual learning of multi-sense embeddings with discrete
  autoencoders.
\newblock In {\em Proceedings of NAACL-HLT\/}. San Diego, California, pages
  1346--1356.

\bibitem[{Tang et~al.(2015)Tang, Qin, and Liu}]{tang2015document}
Duyu Tang, Bing Qin, and Ting Liu. 2015.
\newblock Document modeling with gated recurrent neural network for sentiment
  classification.
\newblock In {\em Porceedings of EMNLP\/}. Lisbon, Portugal, pages 1422--1432.

\bibitem[{Team(2016)}]{2016arXiv160502688short}
Theano~Development Team. 2016.
\newblock Theano: A {Python} framework for fast computation of mathematical
  expressions.
\newblock {\em arXiv e-prints\/} abs/1605.02688.

\bibitem[{Tian et~al.(2014)Tian, Dai, Bian, Gao, Zhang, Chen, and
  Liu}]{tian2014probabilistic}
Fei Tian, Hanjun Dai, Jiang Bian, Bin Gao, Rui Zhang, Enhong Chen, and Tie-Yan
  Liu. 2014.
\newblock A probabilistic model for learning multi-prototype word embeddings.
\newblock In {\em COLING\/}. pages 151--160.

\bibitem[{Tripodi and Pelillo(2017)}]{tripodi2017game}
Rocco Tripodi and Marcello Pelillo. 2017.
\newblock A game-theoretic approach to word sense disambiguation.
\newblock {\em Computational Linguistics\/} 43(1):31--70.

\bibitem[{Tsvetkov et~al.(2015)Tsvetkov, Faruqui, Ling, Lample, and
  Dyer}]{tsvetkov2015evaluation}
Yulia Tsvetkov, Manaal Faruqui, Wang Ling, Guillaume Lample, and Chris Dyer.
  2015.
\newblock Evaluation of word vector representations by subspace alignment.
\newblock In {\em Proceedings of EMNLP (2)\/}. Lisbon, Portugal, pages
  2049--2054.

\bibitem[{Tsvetkov and Wintner(2014)}]{tsvetkov2014identification}
Yulia Tsvetkov and Shuly Wintner. 2014.
\newblock Identification of multiword expressions by combining multiple
  linguistic information sources.
\newblock {\em Computational Linguistics\/} 40(2):449--468.

\bibitem[{Weiss et~al.(2015)Weiss, Alberti, Collins, and
  Petrov}]{weiss2015structured}
David Weiss, Chris Alberti, Michael Collins, and Slav Petrov. 2015.
\newblock Structured training for neural network transition-based parsing.
\newblock In {\em Proceedings of ACL\/}. Beijing, China, pages 323--333.

\bibitem[{Xiao and Cho(2016)}]{XiaoCho:2016}
Yijun Xiao and Kyunghyun Cho. 2016.
\newblock Efficient character-level document classification by combining
  convolution and recurrent layers.
\newblock {\em CoRR\/} abs/1602.00367.

\bibitem[{Yadav et~al.(2017)Yadav, Ekbal, Saha, and
  Bhattacharyya}]{yadav-EtAl:2017:EACLlong}
Shweta Yadav, Asif Ekbal, Sriparna Saha, and Pushpak Bhattacharyya. 2017.
\newblock Entity extraction in biomedical corpora: An approach to evaluate word
  embedding features with pso based feature selection.
\newblock In {\em Proceedings of EACL\/}. Valencia, Spain, pages 1159--1170.

\bibitem[{Yang(1999)}]{yang1999evaluation}
Yiming Yang. 1999.
\newblock An evaluation of statistical approaches to text categorization.
\newblock {\em Information retrieval\/} 1(1-2):69--90.

\bibitem[{Young et~al.(2016)Young, Basile, Kunze, Cabrio, and
  Hawes}]{young2016towards}
Jay Young, Valerio Basile, Lars Kunze, Elena Cabrio, and Nick Hawes. 2016.
\newblock Towards lifelong object learning by integrating situated robot
  perception and semantic web mining.
\newblock In {\em Proceedings of the European Conference on Artificial
  Intelligence conference\/}. The Hague, Netherland, pages 1458--1466.

\bibitem[{Zhong and Ng(2010)}]{ZhongNg:2010}
Zhi Zhong and Hwee~Tou Ng. 2010.
\newblock {It Makes Sense}: A wide-coverage {Word Sense Disambiguation} system
  for free text.
\newblock In {\em Proceedings of the ACL System Demonstrations\/}. Uppsala,
  Sweden, pages 78--83.

\bibitem[{Zou et~al.(2013)Zou, Socher, Cer, and Manning}]{zou2013bilingual}
Will~Y. Zou, Richard Socher, Daniel~M. Cer, and Christopher~D. Manning. 2013.
\newblock Bilingual word embeddings for phrase-based machine translation.
\newblock In {\em Proceedings of EMNLP\/}. Seattle, USA, pages 1393--1398.

\end{thebibliography}
